\documentclass{article}

\usepackage{arxiv}

\usepackage[utf8]{inputenc} 
\usepackage[T1]{fontenc}    
\usepackage{hyperref}       
\usepackage{url}            
\usepackage{booktabs}       
\usepackage{amsfonts}       
\usepackage{nicefrac}       
\usepackage{microtype}      
\usepackage{xcolor}         
\usepackage{lipsum}		
\usepackage{graphicx}
\usepackage{natbib}
\usepackage{doi}
\usepackage{subfigure}
\usepackage{bm}
\usepackage{amsmath}
\usepackage{threeparttable}
\usepackage{multirow}

\title{Gleo-Det: Deep Convolution Feature-Guided Detector with Local Entropy Optimization for Salient Points}

\author{ \href{https://orcid.org/0000-0003-1492-5410}{\includegraphics[scale=0.06]{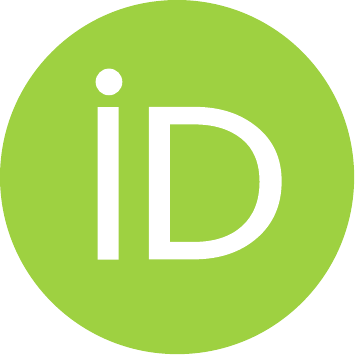}\hspace{1mm}Chao~Li}\\
	School of Artificial Intelligence\\
	Beijing University of Posts and Telecommunications\\
	\texttt{chaoli@bupt.edu.cn} \\
	\And
	\href{https://orcid.org/0000-0001-6473-9187}{\includegraphics[scale=0.06]{orcid.pdf}\hspace{1mm}Yanan~You}\thanks{Corresponding author.}\\
	School of Artificial Intelligence\\
	Beijing University of Posts and Telecommunications\\
	\texttt{youyanan@bupt.edu.cn} \\
	\And
	\href{}{Wenli~Zhou}\\
	School of Artificial Intelligence\\
	Beijing University of Posts and Telecommunications\\
	\texttt{zwl@bupt.edu.cn} \\
}

\renewcommand{\shorttitle}{\textit{arXiv} Template}

\begin{document}
\maketitle

\begin{abstract}
	Feature detection is an important procedure for image matching, where unsupervised feature detection methods are the detection approaches that have been mostly studied recently, including the ones that are based on repeatability requirement to define loss functions, and the ones that attempt to use descriptor matching to drive the optimization of the pipelines. For the former type, mean square error (MSE) is usually used which cannot provide strong constraint for training and can make the model easy to be stuck into the collapsed solution. For the later one, due to the down sampling operation and the expansion of receptive fields, the details can be lost for local descriptors can be lost, making the constraint not fine enough. Considering the issues above, we propose to combine both ideas, which including three aspects. 1) We propose to achieve fine constraint based on the requirement of repeatability while coarse constraint with guidance of deep convolution features.  2) To address the issue that optimization with MSE is limited, entropy-based cost function is utilized, both soft cross-entropy and self-information. 3) With the guidance of convolution features, we define the cost function from both positive and negative sides. Finally, we study the effect of each modification proposed and experiments demonstrate that our method achieves competitive results over the state-of-the-art approaches.
\end{abstract}


\section{Introduction}
Matching salient features across images is an essential premise in many computer vision tasks, such as 3D reconstruction, visual localization and image retrieval. In order to find the local areas that can be matched, one essential premise is to find the salient features with rich semantic information, which is quite crucial in the entire image matching procedure.

Earlier studies mostly concentrate on corner features and blob features, which are of the handcrafted methods based on the expert knowledge. Corner features are usually extracted based on the gradient, intensity or curvature information, among which the first two are more frequently applied, for instance, Harris \citep{harris1988combined}, SUSAN \citep{smith1997susan}, FAST \citep{trajkovic1998fast}, ORB \citep{rublee2011orb} and AGAST \citep{mair2010adaptive}. Different from the corner features, a blob feature generally contains the blob shape information, including scale and orientation. Classical blob features can be generated based on second-order partial derivative and segmentation, such as Laplacian of Gaussian (LoG) \citep{lindeberg1998feature}, difference of Gaussian (DoG) \citep{lowe1999object, lowe2004distinctive}, SIFT \citep{lowe1999object, lowe2004distinctive}, SURF \cite{bay2006surf} and MSER \citep{matas2004robust}.

Intuitively, on the one hand, features with rich semantic information are more than just handcrafted ones, on the other hand, some of the handcrafted features may contain little semantic information. With the application of the convolution neural networks (CNN) \citep{lecun1998gradient} in image processing, learning-based pipelines for feature detection have been studied in recent years. Supervised approaches like LIFT \citep{yi2016lift} and TILDE \citep{verdie2015tilde} select the prominent handcrafted features as positive samples and the smooth areas as the negative samples to train the network. Self-supervised methods involves SuperPoint \citep{detone2018superpoint}, which firstly trains the MagicPoint \citep{detone2018superpoint} based on simulated data set and then generates pseudo labels for further training. Unsupervised pattern is the one that has been studied most, including the ones that are based on repeatability requirement like LF-Net \citep{ono2018lf}, RF-Net \citep{shen2019rf}, UnsuperPoint \citep{christiansen2019unsuperpoint} and Key.Net \citep{barroso2019key}, and the ones that use descriptor matching to drive the optimization of the detector such as D2-Net \citep{dusmanu2019d2}, R2D2 \citep{revaud2019r2d2}, and ASLFeat \citep{luo2020aslfeat}.

However, for the methods that leverage the feature repeatability and use MSE cost function when training, including LF-Net \citep{ono2018lf}, RF-Net \citep{shen2019rf}, UnsuperPoint \citep{christiansen2019unsuperpoint} and Key.Net \citep{barroso2019key}, constraint is weak and the model can be stuck into collapsed solution, making convergence difficult. While for the approaches that use local features to drive the learning of feature detector and train the descriptor and detector jointly, such as D2-Net \citep{dusmanu2019d2}, R2D2 \citep{revaud2019r2d2} and ASLFeat \citep{luo2020aslfeat}, the constraint for detector is not fine enough due to down sampling operation and expansion of the receptive fields.

For the issue mentioned above, our contributions can be illustrated from three aspects. First, fine constraint is achieved via repeatability cost function while coarse constraint is realized with the guidance of convolution features. Second, to address the issue that optimization via MSE is limited for detection learning, we trained the detector with entropy-based loss function, making convergence easier. Third, we define cost function from both positive and negative sides under the guidance of convolution features, which aims to reject the key points with little semantic information. Eventually, our method shows good repeatability and mean matching accuracy, which illustrates that our method is not only precise for feature localization but also robust for matching.

\section{Related Works}
Detected features are special structures which represent certain semantic information in images, including corner features, blob features, edges and morphological region features. The most popular features used for image matching are key points. Good key point detection algorithm should be (i) stable and (ii) matchable. In other words, on the one hand, key points detected should be invariant to both illumination changes and spatial transform. on the other hand, they should also be rich in semantic information so that can be matched. 

\subsection{Handcrafted features}
The most common handcrafted features used are corner features and blob features. A corner feature is a the cross point of two lines, which can be depicted in the form of "L", "T" and "X". To find out the corner features in an image, one approach is to leverage the gradient information, like Harris \citep{harris1988combined}, and the other is to detect via intensity information directly, including SUSAN \citep{smith1997susan}, FAST \citep{trajkovic1998fast} and ORB \citep{rublee2011orb}. On the contrary, a blob feature is regarded as a closed region with not only the feature location information $(x,y)$ but also the blob shape information $(s, \theta)$, indicating scale and orientation. Typical blob features involve classic Laplacian of LoG \citep{lindeberg1998feature}, DoG \citep{lowe1999object, lowe2004distinctive}, SIFT \citep{lowe1999object, lowe2004distinctive}, etc.

Harris \citep{harris1988combined} firstly calculates the auto-correlation matrix with the gradient information for each location, and then finds the directions of the fastest and lowest grey-value changes to determine whether it is the key point. SUSAN \citep{smith1997susan} assumes a pixel as a key point when the intensity values of enough pixels in its neighboring area are different, and the distance between the centroid of the neighboring area and the pixel is far enough. FAST \citep{trajkovic1998fast} takes comparison with intensity values of each pixel on a circle, and selects corner features using a machine learning approach trained on a large number of similar scene images. ORB \citep{rublee2011orb} uses the Harris \citep{harris1988combined} response to select FAST \citep{trajkovic1998fast} key points as the final detected features.

Generally, the corner detection algorithms mentioned above are rotation-invariant but not scale-invariant without taking sale into consideration. Nevertheless, blob features are quite differnt. LoG \citep{lindeberg1998feature} determines blob features through detecting local extremums on images convolved by scale-normalized Laplacian of Gaussian under different scales defined with the standard deviation $\sigma$. DoG \citep{lowe1999object, lowe2004distinctive} is an improved version of scale-normalized LoG as they are approximately equal, which is more efficient. SIFT \citep{lowe1999object, lowe2004distinctive} firstly constructs the DoG scale space and detects local extremums to obtain feature locations as well as scales. Subsequently, through the gradient histogram it calculates the directions for each key point. SIFT is a stable feature detection algorithm that has been widespread used in numerous tasks. Compared to the corner features above, blob features are usually both rotation-invariant as well as scale-invariant via using different $\sigma$s as scales. 

\subsection{Learning-based features}
One issue for conventional handcrafted features like corners and blobs is that these features might not contain enough semantic information for matching in some cases. And numerous studies attempt to apply deep learning approaches to feature detection, in both supervised, self-supervised and unsupervised manners.

Supervised detectors include TILDE \citep{verdie2015tilde} and LIFT \citep{yi2016lift}. TILDE \citep{verdie2015tilde} generates the data set using patches with confident SIFT key points as positive samples and the ones far from key points as negative samples, and the regressor is a piece-wise linear function expressed via Generalized Hinging Hyperplanes (GHH). LIFT \citep{yi2016lift} takes an inverse order during the training process, first the local descriptor, then the orientation estimator and finally the detector. The orientation estimator is trained under constraints of the descriptors gained before. The detector is also a linear function of GHH, but adds one more constraint based on the orientation estimator and descriptor. One typical self-supervised method is SuperPoint \citep{detone2018superpoint}. SuperPoint firstly generates simulated data like points and line segments with key point labels to train the MagicPoint, and the model is used to generate pseudo labels via homography adaption  subsequently.

For the unsupervised methods, one idea is to use the repeatability of key points, like LF-Net \citep{ono2018lf}, RF-Net \citep{shen2019rf}, UnsuperPoint \citep{christiansen2019unsuperpoint} and Key.Net \citep{barroso2019key}, another idea is to utilize the local descriptors to drive the training process of the detector, like D2-Net \citep{dusmanu2019d2}, R2D2 \citep{revaud2019r2d2} and ASLFeat \citep{luo2020aslfeat}. Key.Net \citep{barroso2019key} proposes the index proposal (IP) layer based on softmax operation for location regression, and defines the MSE loss of correspondence location error weighted by corresponding responses. While LF-Net \citep{ono2018lf}, RF-Net \citep{shen2019rf} and UnsuperPoint \citep{christiansen2019unsuperpoint} use the MSE of responses as loss function instead. Actually, these approaches are usually not easy to train due to lack of strong constraints. Moreover, for training with MSE of responses, the detector is easy to be stuck into collapsed solution where all scores corresponding to each pixel are 0. Approaches like D2-Net \citep{dusmanu2019d2}, R2D2 \citep{revaud2019r2d2} and ASLFeat \citep{luo2020aslfeat} jointly train the descriptor and detector, where the detector and the descriptor are optimized via one shared loss function. Constructed based on convolution features with a shared loss function, from which the detector can derive semantic information. However, the information cannot guide fine location due to the expansion of receptive fields and loss of detailed information in the local features.

\begin{figure}[htpb]
	\centering
	\subfigure[]{
		\label{Fig1.sub.1}
		\includegraphics[height=0.48\textwidth]{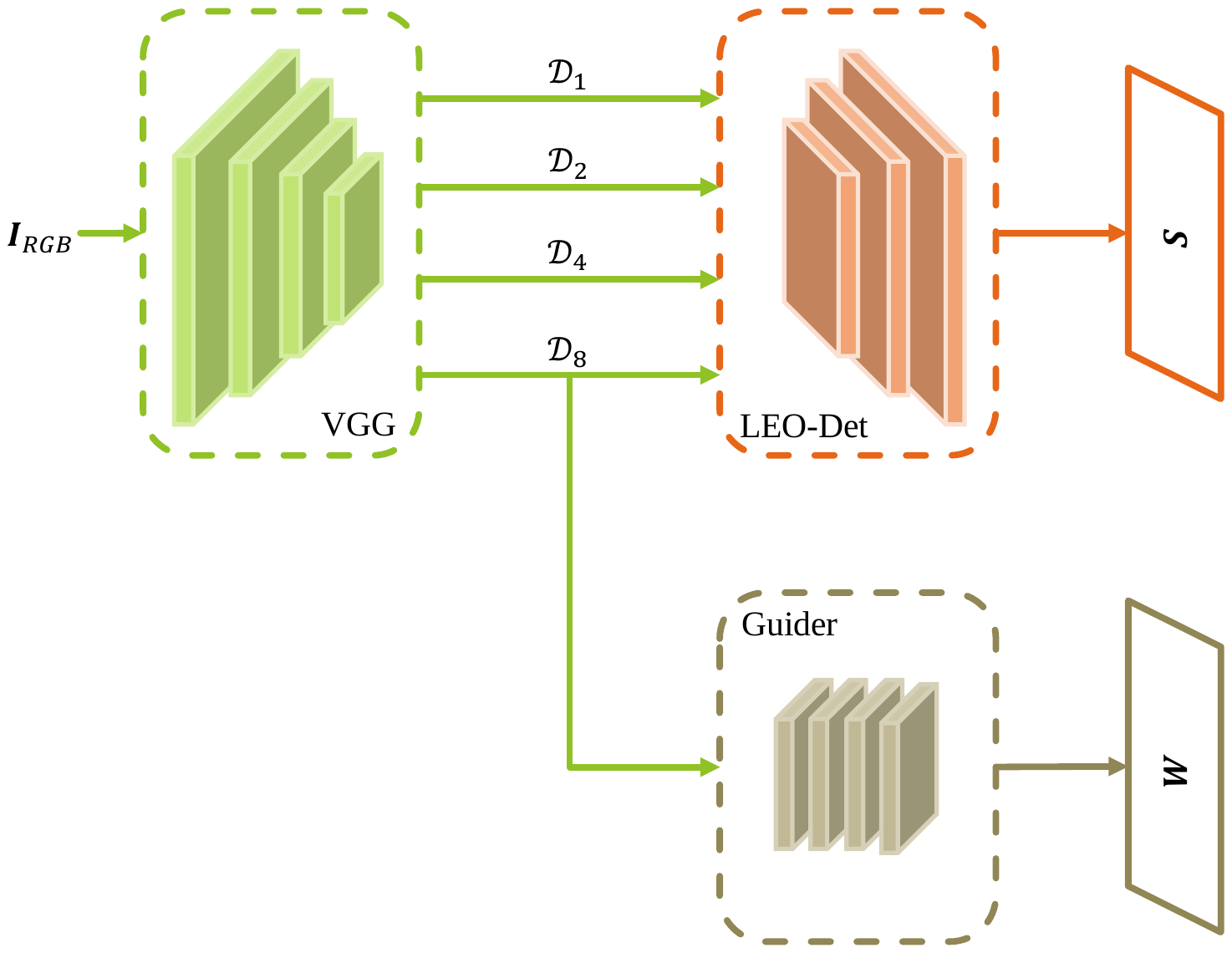}}\hspace{30pt}
	\subfigure[]{
		\label{Fig1.sub.2}
		\includegraphics[height=0.48\textwidth]{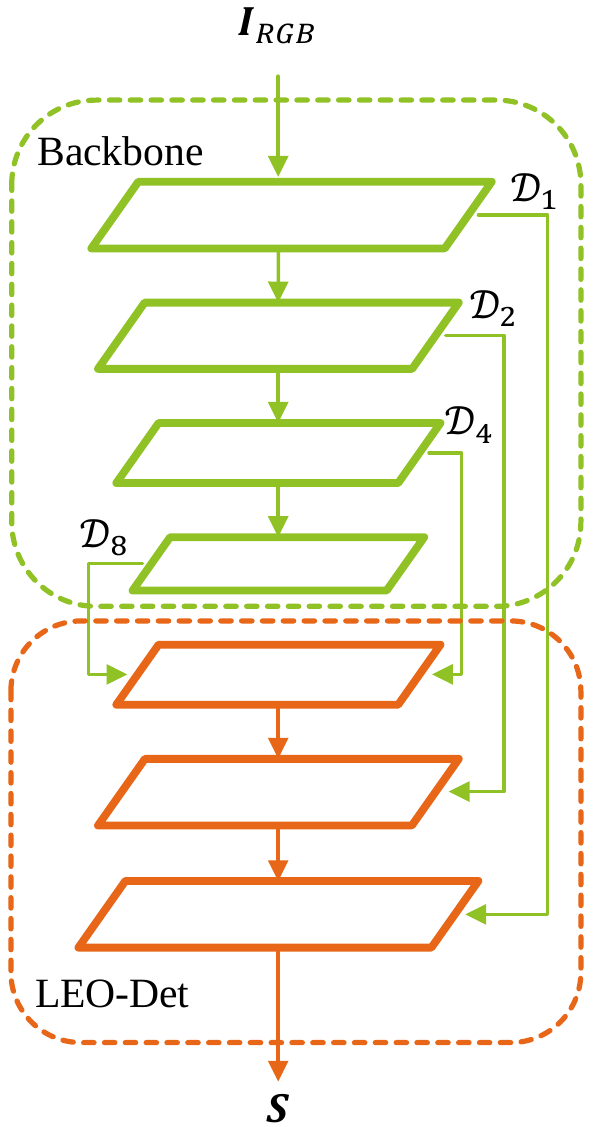}}
	\caption{\small{(a) shows the proposed pipeline to get the score map $\bm{S}$ and the weight map $\bm{W}$, where $\mathcal{D}_k, k=1,2,4,8$ refers to the feature maps of different convolutional layers and k is the downsampling rate. For score map $\bm{S}$, ReLU is used to make all elements positive and for the weight map $\bm{W}$, sigmoid is adopted to ensure all values range from 0 to 1. (b) shows the UNet structures of the LEO-Det in this paper.}}
	\label{Fig1.main}
\end{figure}

\section{Approaches}
\subsection{Detector Structure}
The structure of our guided local entropy optimal detector (GLEO-Det), is shown in Fig. 1 (a), consisting of three parts which are the backbone, the guider as well as the LEO-Det. In this paper, we use VGG-16 \citep{simonyan2014very} as the backbone, where the former 4 blocks are used and the output feature map of each block is the input of the proposed LEO-Det. As for the specific architecture of the LEO-Det, we try UNet \citep{ronneberger2015u} structure as shown in Fig. 1(b). In the Guider block, an extra convolution block is adopted to give a weight map, where Sigmoid operation is followed to ensure the weight map ranges form 0 to 1, indicating the importance of each output feature of the backbone. In the backbone of VGG-16, no normalization operation is needed, while in the Guider and the LEO-Det parts, the instance normalization \citep{ulyanov2016instance} is applied. 

\subsection{Local Entropy-based (LE) Loss}
Current unsupervised pipelines usually optimize repeatability via MSE loss, which cannot provide enough constraint during the training process, and can be easily stuck into collapsed solution when MSE of responses are directly adopted. To address the issue, we use the local soft cross-entropy loss function for further learning when the response distributions for corresponding locations are the same. Using this approach, we only need one epoch with 2000 iterations for the detector training process, where one image is used in each iteration, showing the efficiency of our method.

Suppose the response maps of images $\bm{I}_1$ and $\bm{I}_2$ with overlapped areas are denoted as $\bm{S}_1$ and $\bm{S}_2$, where the ground true homography from $\bm{I}_1$ to $\bm{I}_2$ is $\bm{H}$ and the spatial mapping is $t$. 

Assume that the size of both images are $H\times{W}=8w\times8h$. In this paper, like SuperPoint, we assume adopt the $8\times{8}$ grid as a sample for loss function. With the ground true spatial mapping $t$ from $\bm{I}_1$ to $\bm{I}_2$, we have the aligned response map pairs $(\bm{S}_1, t^{-1}(\bm{S}_2))$ and $(t(\bm{S}_1), \bm{S}_2)$. Then reshape them into 3D tensors denoted as $(\mathcal{S}_1, \mathcal{S}_2^t)$ and $(\mathcal{S}_1^t, \mathcal{S}_2)$, which are all of $64\times{w}\times{h}$. In this paper, we use the location with the highest score in each grid to select training samples. The feature point index matrices of $\bm{I}_1$ and $\bm{I}_2$ are written as,
\begin{equation}
\bm{K}_{1ij} = \mathop{\arg\max}\limits_{1\leq{k}\leq{64}}{\mathcal{S}_{1kij}},\;\bm{K}_{2ij} = \mathop{\arg\max}\limits_{1\leq{k}\leq{64}}{\mathcal{S}_{2kij}}
\end{equation}
Then we select the indices in $\mathcal{S}_1$ and $\mathcal{S}_2$ to build set $A$ defined as follow,
\begin{equation}
\begin{aligned}
x_{ij} &= 8(j-1) + (\bm{K}_{1ij}\mod{8}),\quad y_{ij} = 8(i-1) +\lfloor \bm{K}_{1ij}/8\rfloor\\
p_{ij} &= 8(j-1) + (\bm{K}_{2ij}\mod{8}),\quad q_{ij} = 8(i-1) +\lfloor \bm{K}_{2ij}/8\rfloor\\
A&=\{(a,b,c,d)
\mid{\lfloor{f(\bm{H},[x_{ab},y_{ab}]^T)/{8}}\rfloor=[c,d]^T \wedge{\lfloor{f(\bm{H},[p_{cd},q_{cd}]^T)/{8}}\rfloor=[a,b]^T} }
\}
\end{aligned}
\end{equation}
Let $[x',y',z'] = \bm{H}[x,y,1]^T$, then $f$ is defined as,
\begin{equation}
\begin{aligned}
f(\bm{H},[x,y]^T) = [x'/z', y'/z']^T
\end{aligned}
\end{equation}
To define the loss function, we write the slices of the tensors as,
\begin{equation}
\begin{aligned}
\bm{u}_{ij} = sm(\mathcal{S}_{1:ij}),\, \bm{u}'_{ij} = sm(\mathcal{S}^t_{1:ij})\\
\bm{v}_{ij} = sm(\mathcal{S}_{2:ij}),\; \bm{v}'_{ij} = sm(\mathcal{S}^t_{2:ij})
\end{aligned}
\end{equation}
where $sm$ refers to softmax operation and each element can be seen as the probability of the corresponding location to be a key point. The local similarity matrices can be defined in the form of cross-entropy as follow.
\begin{equation}
\begin{aligned}
\bm{E}_{1ij} &= -\sum_{k}(\bm{u}_{ijk}\log{\bm{v}'_{ijk}} + \bm{v}'_{ijk}\log{\bm{u}_{ijk}})\\
\bm{E}_{2ij} &= -\sum_{k}(\bm{v}_{ijk}\log{\bm{u}'_{ijk}} - \bm{u}'_{ijk}\log{\bm{v}_{ijk}})
\end{aligned}
\end{equation}
For each $\bm{x}=(a,b,c,d)\in{A}$, let $\bm{E}_{1\bm{x}}=\bm{E}_{1ab}$, $\bm{E}_{2\bm{x}}=\bm{E}_{2cd}$, and the LE loss is define below,
\begin{equation}
\mathcal{L}_{le} = \frac{1}{4|A|}\sum_{\bm{x}\in{A}}(\bm{E}_{1\bm{x}}+\bm{E}_{2\bm{x}})
\end{equation}

\subsection{Guided Local Entropy-based (GLE) Loss}
Guider is to point out the significant areas with salient features and smooth areas in the image and assign ballots for entropy-based loss of each local grid mentioned above. In addition, with the Guider, we can further force the energy of grids with key points to gather and that of smooth grids to scatter on the other side.

Suppose the weight maps output by Guider is denoted as $\bm{W}_1$ and $\bm{W}_2$ respectively. For each $\bm{x}=(a,b,c,d)\in{A}$, we let $\bm{W}_{1\bm{x}}=\bm{W}_{1ab}$, $\bm{W}_{2\bm{x}}=\bm{W}_{2cd}$, $\bm{W}_{\bm{x}}=\bm{W}_{1\bm{x}}\bm{W}_{2\bm{x}}$. With the weight maps, we can rewrite the LE loss as follow,
\begin{equation}
\mathcal{L}_{le}^W = \frac{1}{4\sum_{\bm{x}\in{A}}\bm{W}_{\bm{x}}}\sum_{\bm{x}\in{A}}(\bm{E}_{1\bm{x}}+\bm{E}_{2\bm{x}})\bm{W}_{\bm{x}}
\end{equation}

For further improvement of the repeatability under more strict requirements, we add the term of certainty via self-information of each grid. In other words, grids with high weight should be more confident in key point location. Based on this, we use certainty matrices to depict the confidence of each grid.
\begin{equation}
\begin{aligned}
\bm{C}_{1ij} &= -\sum_{k}(\bm{u}_{ijk}\log{\bm{u}_{ijk}} + \bm{v}'_{ijk}\log{\bm{v}'_{ijk}})\\
\bm{C}_{2ij} &= -\sum_{k}(\bm{v}_{ijk}\log{\bm{v}_{ijk}} - \bm{u}'_{ijk}\log{\bm{u}'_{ijk}})
\end{aligned}
\end{equation}
The local certainty loss, the other term of GLE loss, is
\begin{equation}
\mathcal{L}_{lc}^W = \frac{1}{4\sum_{\bm{x}\in{A}}\bm{W}_{\bm{x}}}\sum_{\bm{x}\in{A}}(\bm{C}_{1\bm{x}}+\bm{C}_{2\bm{x}})\bm{W}_{\bm{x}}-\frac{1}{4\sum_{\bm{x}\in{A}}(1-\bm{W}_{\bm{x}})}\sum_{\bm{x}\in{A}}(\bm{C}_{1\bm{x}}+\bm{C}_{2\bm{x}})(1-\bm{W}_{\bm{x}})
\end{equation}
where for each $\bm{x}=(a,b,c,d)\in{A}$, $\bm{C}_{1\bm{x}}=\bm{C}_{1ab}$, $\bm{C}_{2\bm{x}}=\bm{C}_{2cd}$. $\mathcal{L}_{lc}^W$ not only requires the grids with high weights to be more certain for key point location, but also requires girds with low weights to be ambiguous on the other side. And the guided local entropy loss is a combination of both parts, which is,
\begin{equation}
\mathcal{L}_{gle} =  \mathcal{L}_{le}^W + \mathcal{L}_{lc}^W
\end{equation}

\subsection{Inference} 
In this paper, we train our model using GLE loss function. With the score map $\bm{S}$ output by the trained detector, we use the weight map $\bm{W}$ of the Guider to remove the features that are not suitable for matching. Specifically, the weighted score map is written as follow,
\begin{equation}
\bm{S_{w}} =  Int(\bm{W})\odot{\bm{S}}
\end{equation}
where $Int$ refers to the bilinear interpolation operation and $\odot$ is Hardmard product. In the validation stage, we choose the top $k$ locations with the highest responses as key points for matching. In this paper, we take $k=3000$.

\subsection{Data set}
In this paper, we adopt the Random Web Images in R2D2 to train our model, which contains 3125 images. To train the detector, random projection transform as well as color and contrast adjustments are used to construct image pairs with augmentation. Besides, the classic HPatches \citep{balntas2017hpatches} data set is used for validation. The data set contains 116 sequences, where each contains 6 images with the homography known between each other. HPatches data set involves two types of changes, including 59 sequences with \textit{viewpoint} changes and 57 sequences with \textit{illumination} changes.

\section{Experiment}
\begin{figure}[htpb]
	\centering
	\subfigure[Original Image.]{
		\label{Fig2.sub.1}
		\includegraphics[width=0.2\textwidth]{image.jpg}}
	\subfigure[RegMSE.]{
		\label{Fig2.sub.2}
		\includegraphics[width=0.2\textwidth]{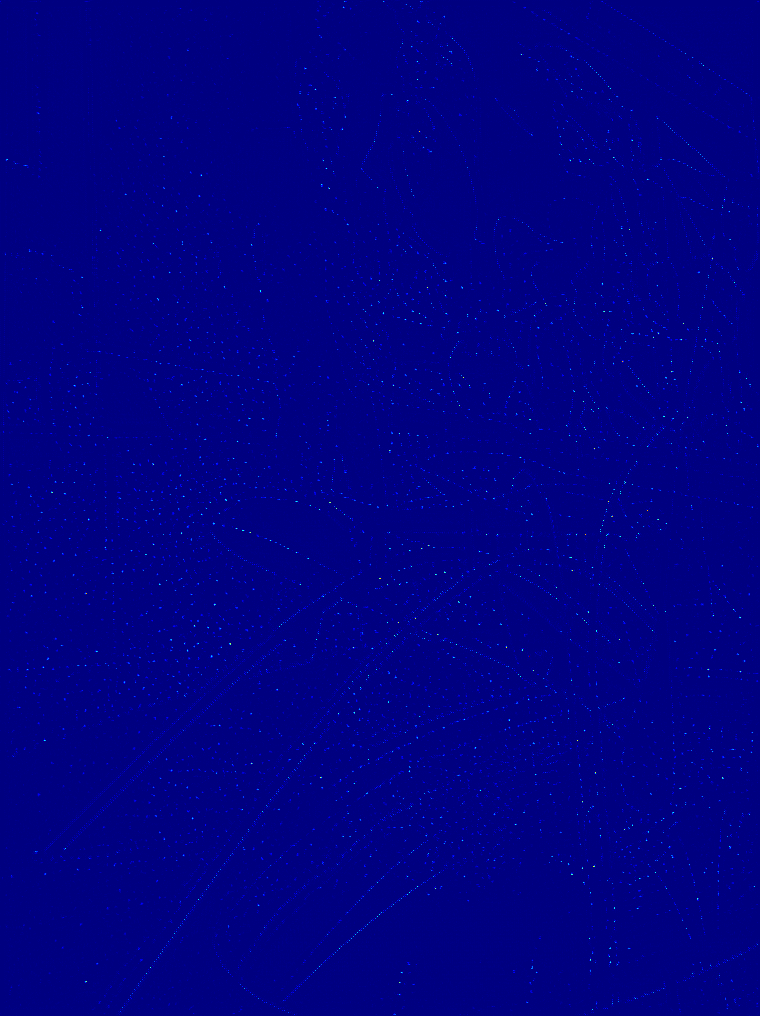}}
	\subfigure[LE.]{
		\label{Fig2.sub.3}
		\includegraphics[width=0.2\textwidth]{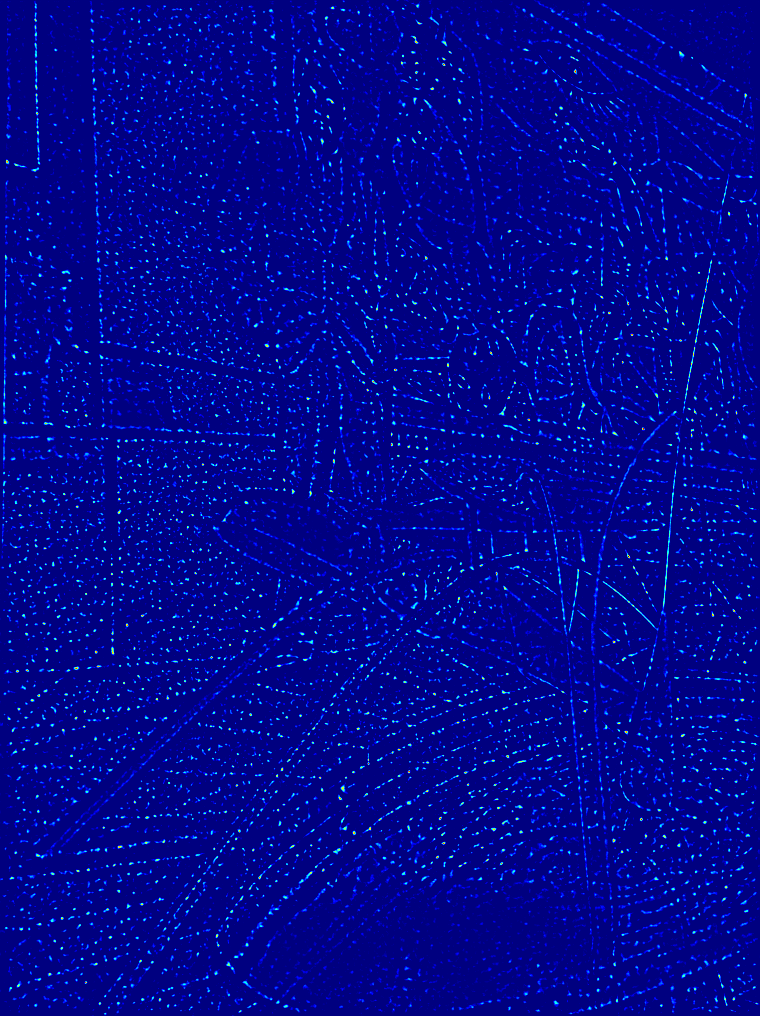}}
	\subfigure[GLE.]{
		\label{Fig2.sub.4}
		\includegraphics[width=0.2\textwidth]{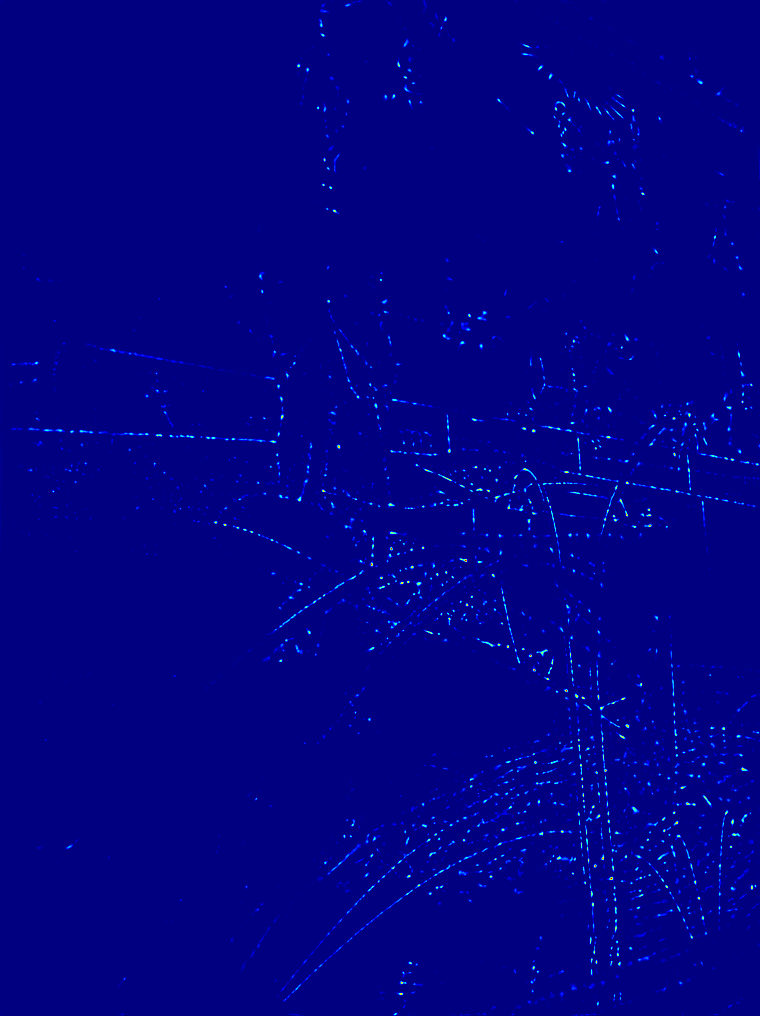}}
	\subfigure[Original Image.]{
		\label{Fig2.sub.5}
		\includegraphics[width=0.2\textwidth]{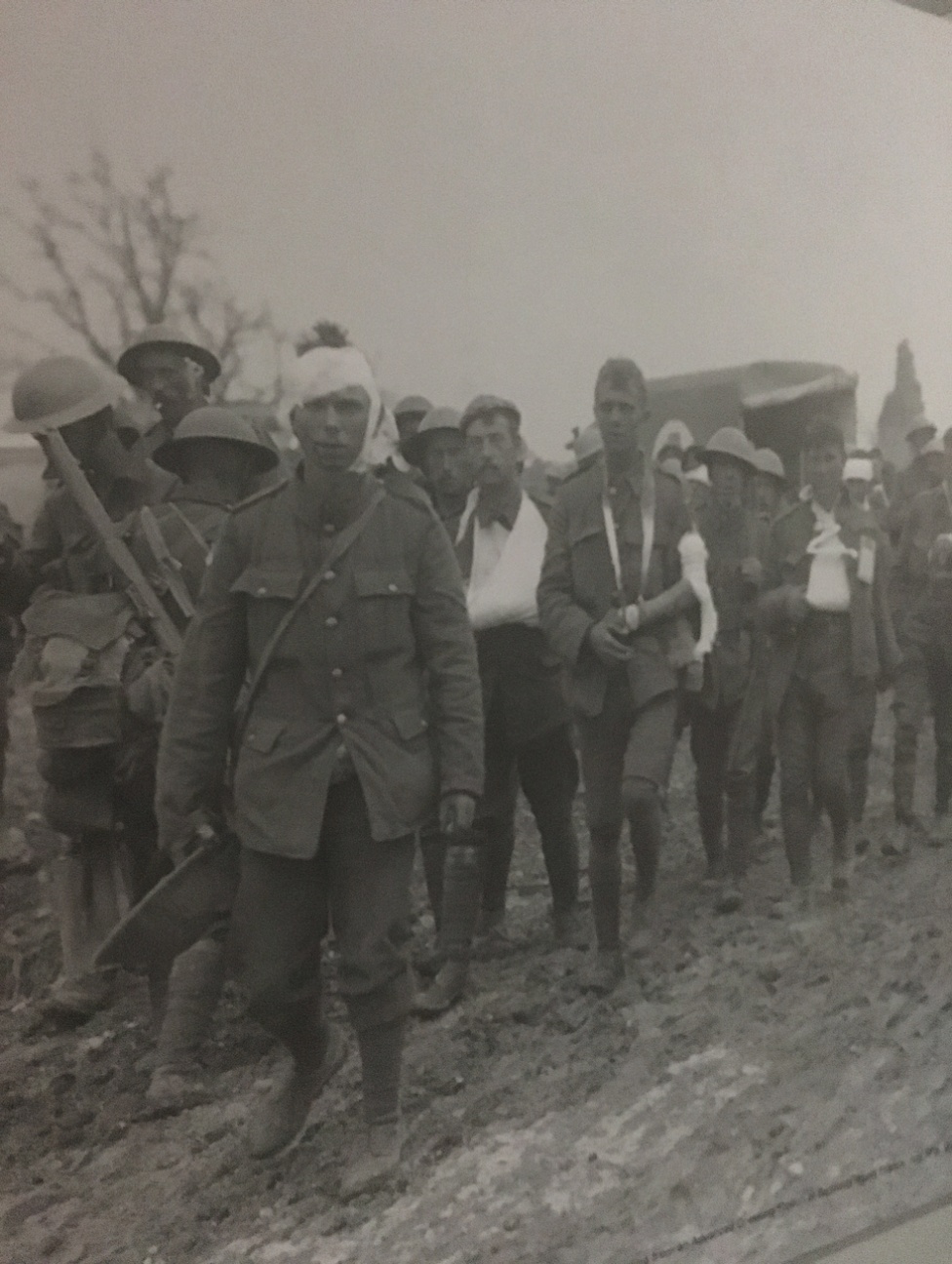}}
	\subfigure[RegMSE.]{
		\label{Fig2.sub.6}
		\includegraphics[width=0.2\textwidth]{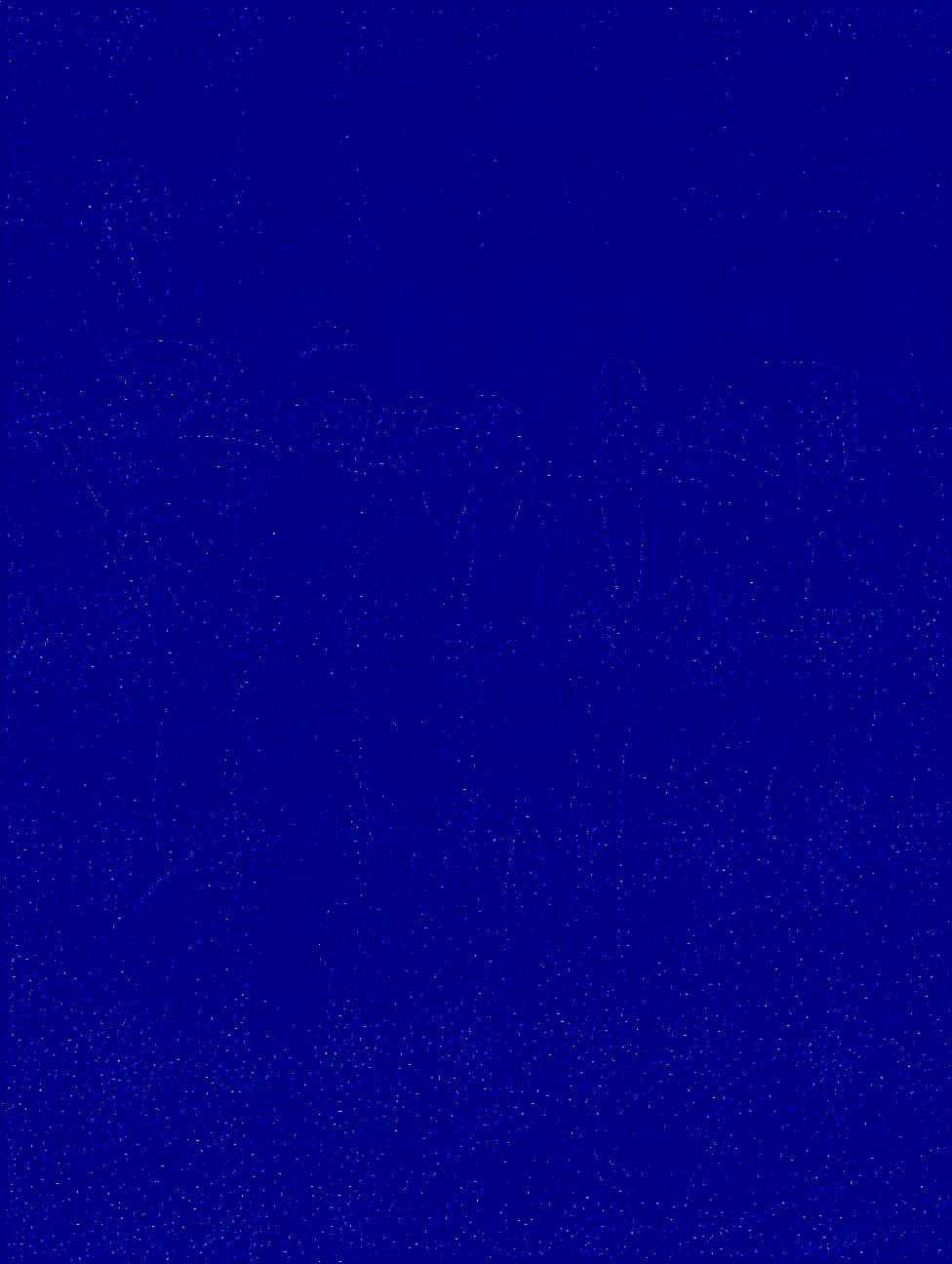}}
	\subfigure[LE.]{
		\label{Fig2.sub.7}
		\includegraphics[width=0.2\textwidth]{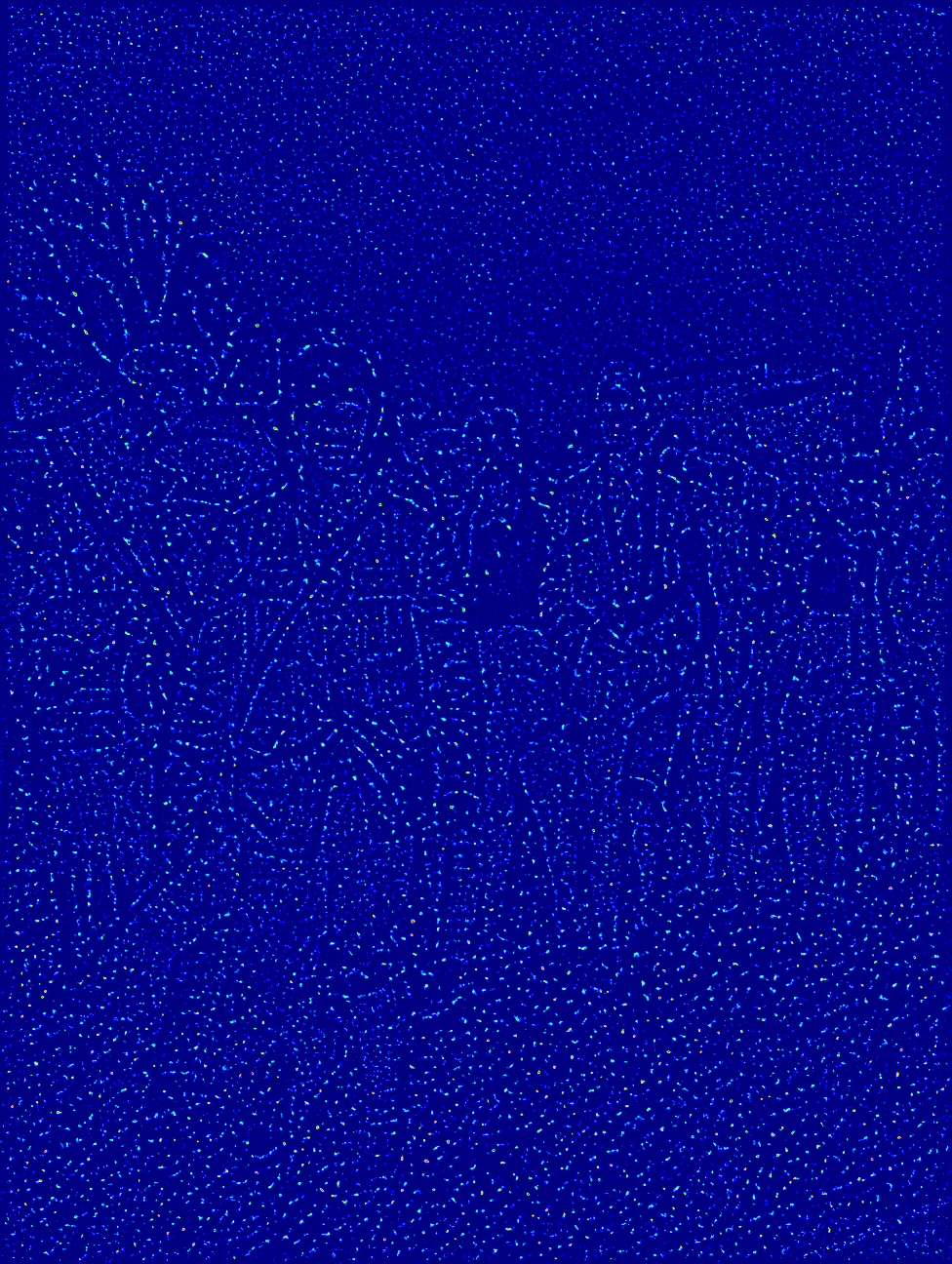}}
	\subfigure[GLE.]{
		\label{Fig2.sub.8}
		\includegraphics[width=0.2\textwidth]{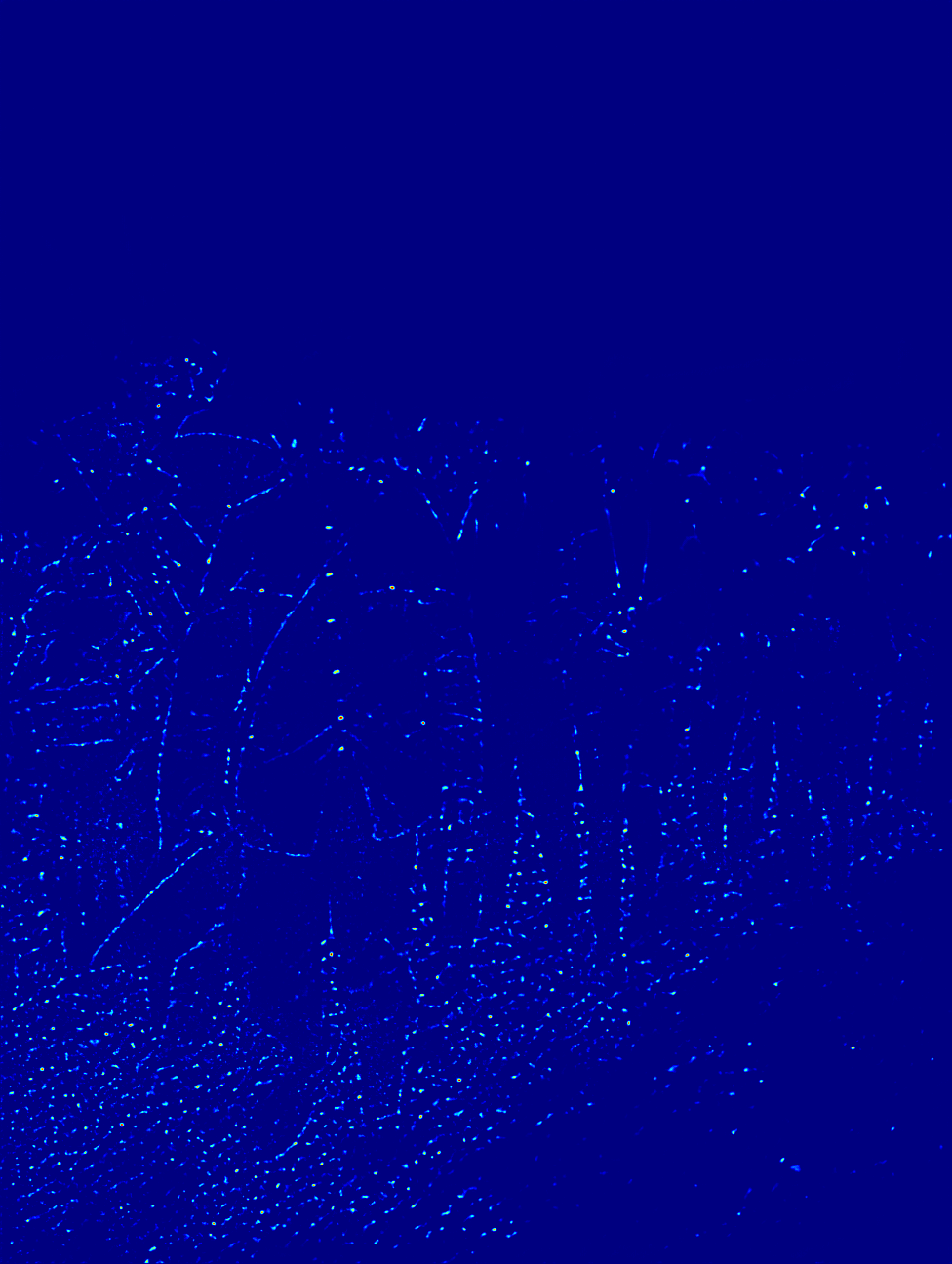}}
	\subfigure[Original Image.]{
		\label{Fig2.sub.9}
		\includegraphics[width=0.2\textwidth]{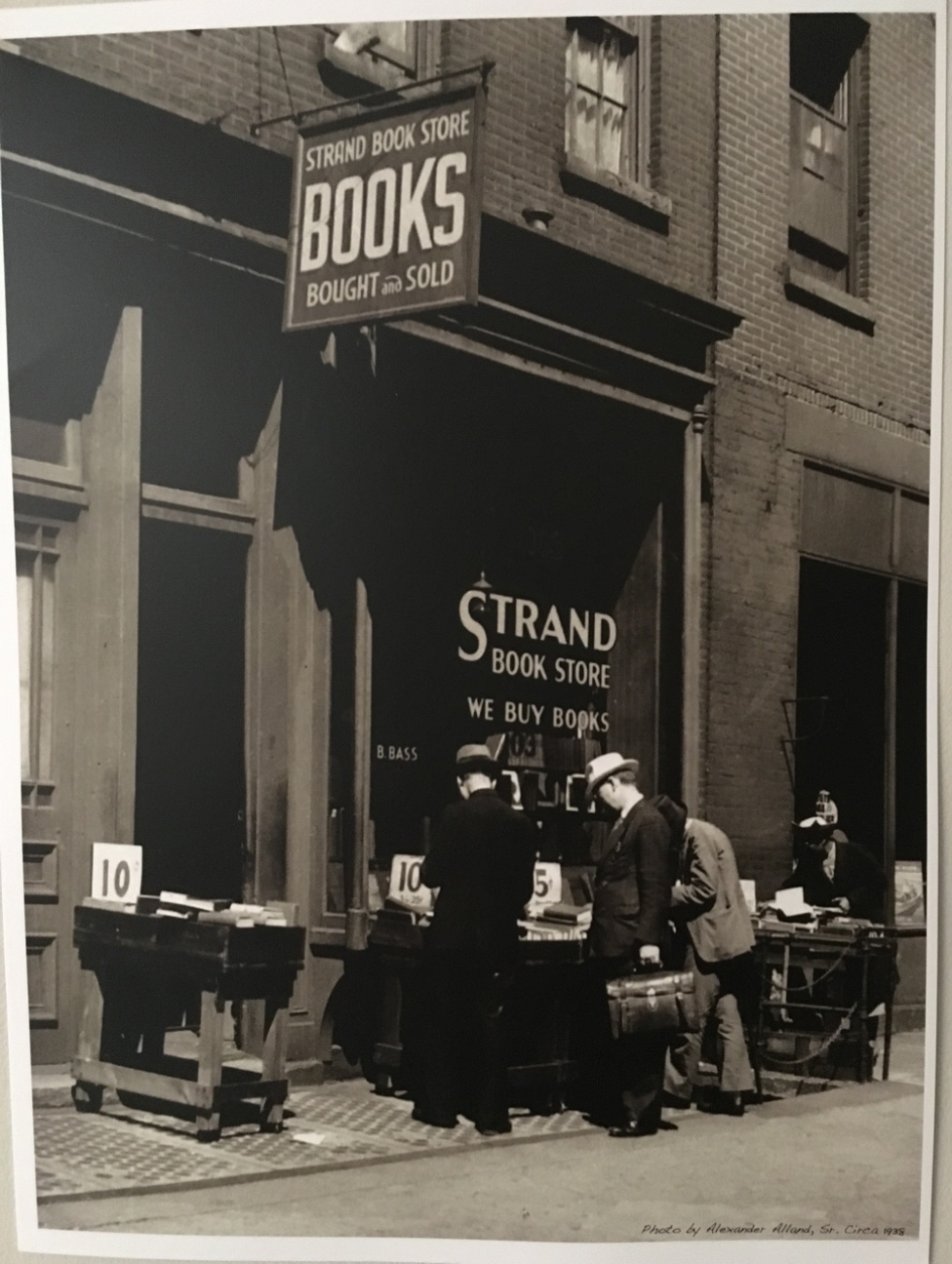}}
	\subfigure[RegMSE.]{
		\label{Fig2.sub.10}
		\includegraphics[width=0.2\textwidth]{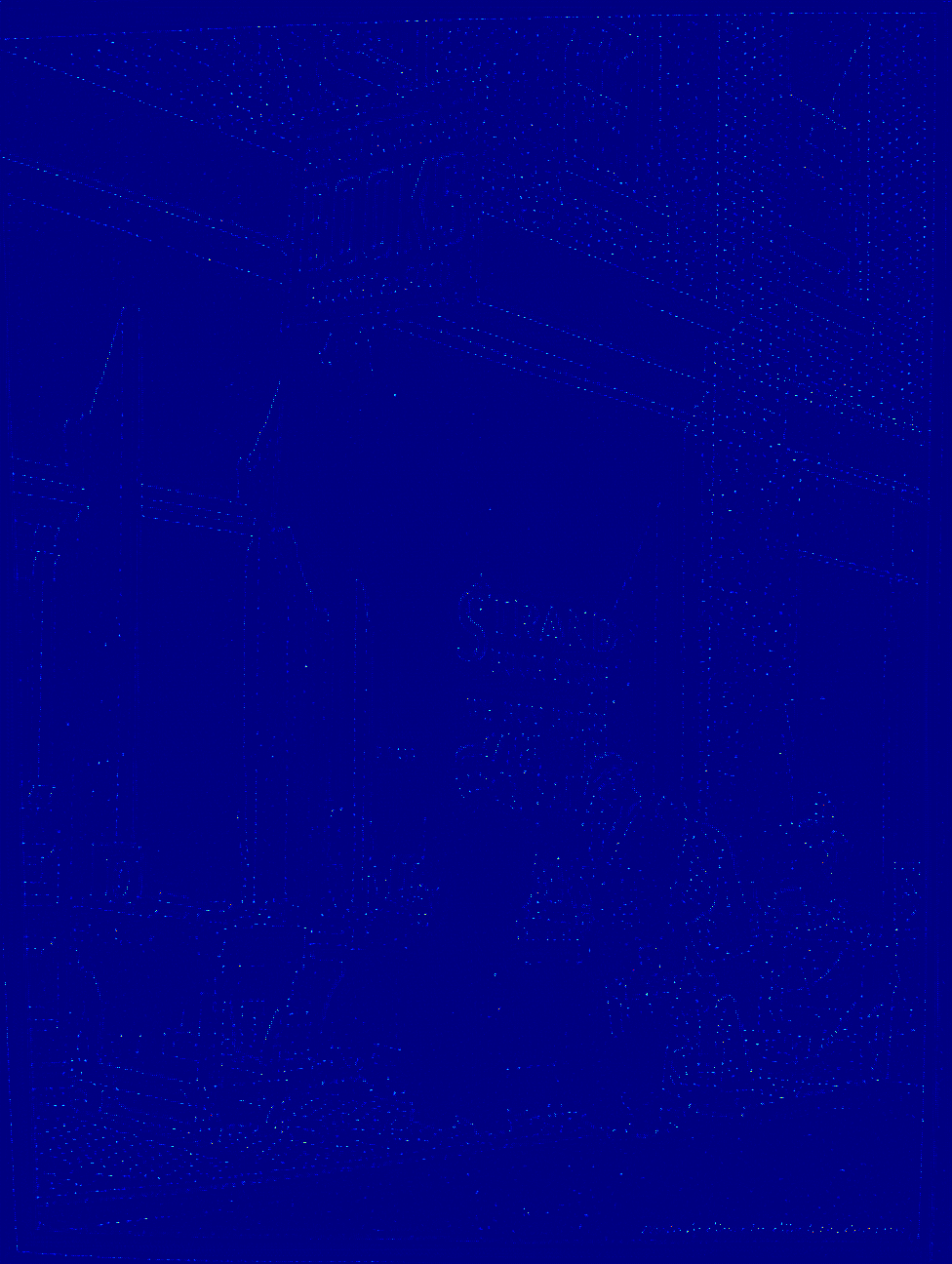}}
	\subfigure[LE.]{
		\label{Fig2.sub.11}
		\includegraphics[width=0.2\textwidth]{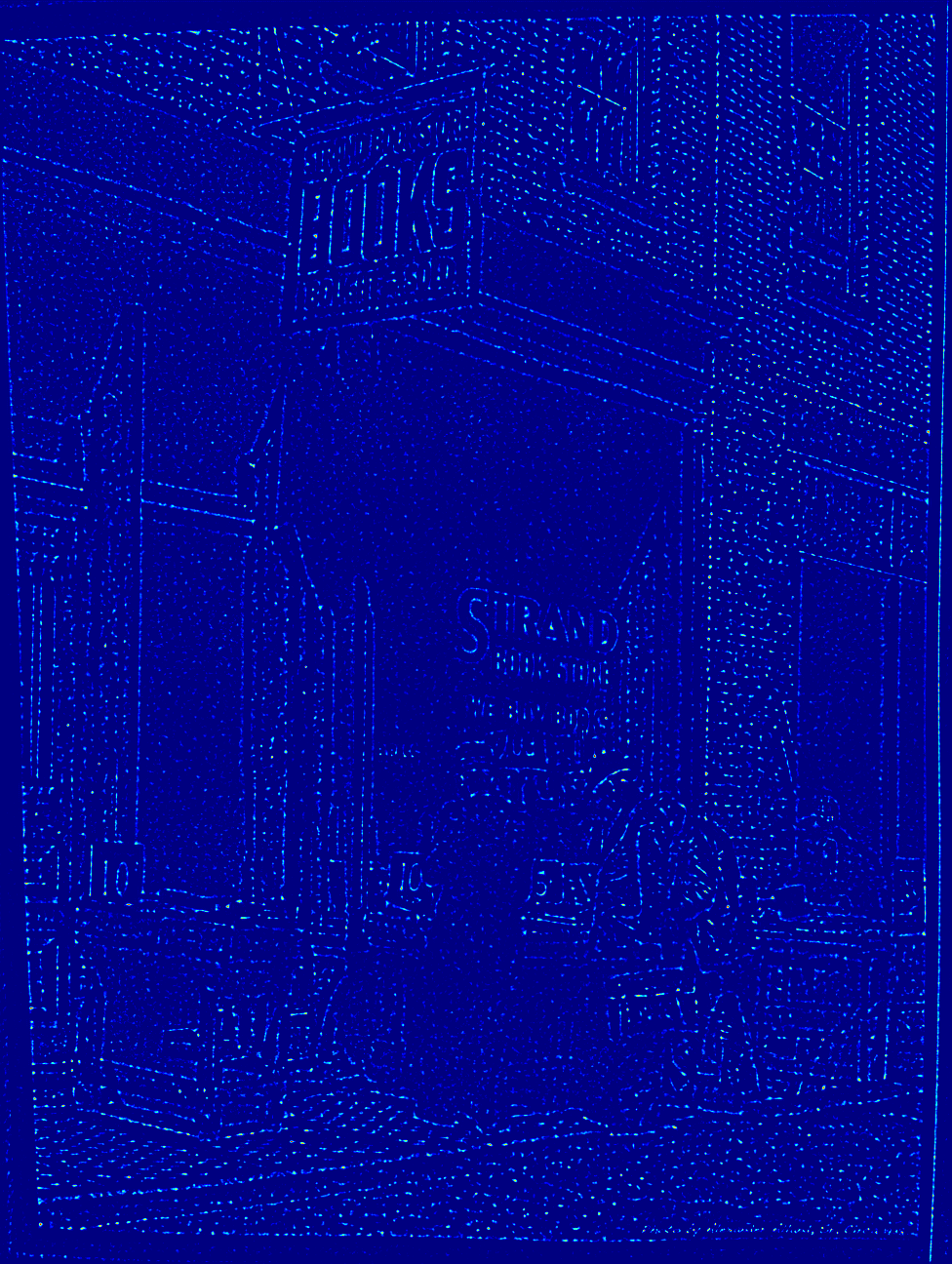}}
	\subfigure[GLE.]{
		\label{Fig2.sub.12}
		\includegraphics[width=0.2\textwidth]{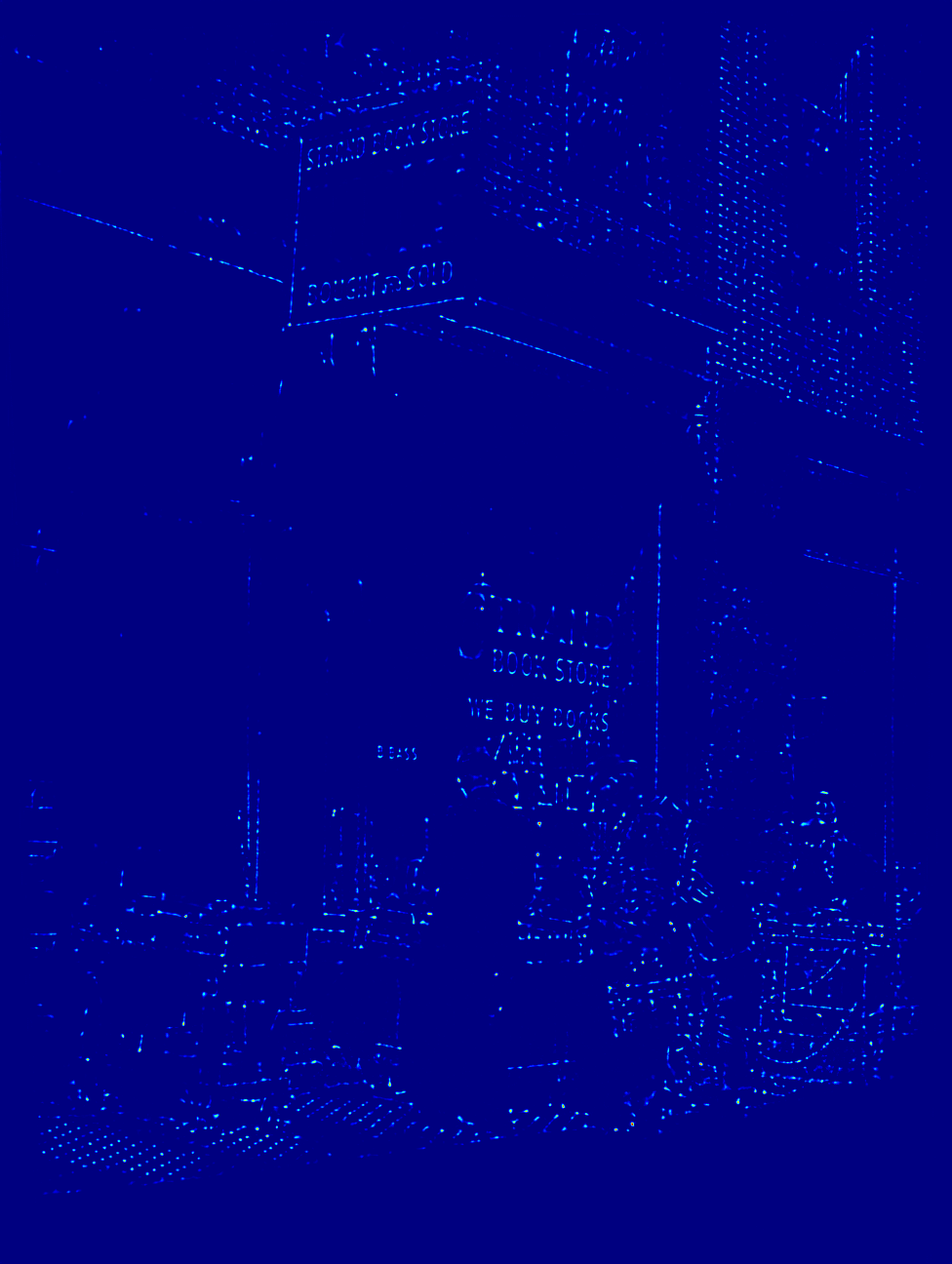}}
	\caption{\small{Score maps using different methods in Tab. 1, and we can see that the Guider module is essential in rejecting points with little semantic information.}}
	\label{Fig2.main}
\end{figure}

\begin{table}[htpb]
	\centering
	\caption{\small{Ablation studies. The table shows the repeatability(Rep) and mean matching accuracy(MMA) at precision threshold $\epsilon=1$. For evaluation with MMA, HardNet++(HN++) is used to construct descriptors. We denote the detector structure as $F$ and test it trained with RegMSE, LE and GLE loss function respectively.}}
	\begin{threeparttable}
		\begin{tabular}{ccccccc}
			\toprule
			\multirow{2}{*}{} & \multicolumn{2}{c}{HPatches \textit{illu}} & \multicolumn{2}{c}{HPatches \textit{view}} &  \multicolumn{2}{c}{ \textit{Overall}}\\ 
			& \%rep@1 & \%mma@1 & \%rep@1 & \%mma@1 & \%rep@1 & \%mma@1\\
			\midrule
			\multicolumn{1}{c}{$F$+RegMSE+HN++}           &37.16  &59.03  &19.96  &25.34  &28.41  &41.56 \\
			\multicolumn{1}{c}{$F$+LE+HN++}       &42.56  &61.93  &30.27  &31.82  &36.31  &46.32\\
			\multicolumn{1}{c}{$F$+GLE+HN++}           &\textbf{48.43}  &\textbf{69.07}  &\textbf{42.84}  &\textbf{37.73}  &\textbf{45.58}  &\textbf{52.82}\\
			\bottomrule
		\end{tabular}
	\end{threeparttable}
\end{table}

\subsection{Ablation studies}
In this part, we attempt to validate the effect of LE Loss and the GLE loss. To show that the entropy-based loss is more suitable for detector training, as a control group, we take the MSE loss to optimize the response map, and use the distance between the highest response and average response of the $8\times{8}$ grid for regularization, which we name as regularized MSE (RegMSE). Furthermore, to verify that the Guider module is helpful for removing features without rich semantic information, we compare the performance of the detectors trained with $\mathcal{L}_{le}$ in Eq. (6) and $\mathcal{L}_{gle}$ in Eq. (10) respectively. For evaluation, we test the repeatability and 
mean matching accuracy (MMA) at error threshold $\epsilon=1$ which is a strict requirement on the HPatches with illumination changes as well as viewpoint changes. The results are shown in Tab. 1. In addition, we also show the score maps of one example using the above three methods in Fig. 2.

From Tab. 1 and the second column of Fig. 2, it can be seen that regularization does help for inhibiting possible collapsed solution. However, using local entropy-based loss function without Guider still performs better when precision requirement is quite strict, and this tells that entropy-based optimization is more suitable for detector training. Moreover, from Tab. 1, we can see that with the Guider block, training with GLE loss is better than only with LE loss and RegMSE. Combining with Fig. 2, we believe that is because the Guider module can help to reject the features with little semantic information.

\begin{table*}[htpb]
	\centering
	\caption{\small{The table shows the repeatability of our method and other methods including typical handcrafted features, self-supervised learning based ones and unsupervised leaning based ones. And we test the cases with the precision threshold $\epsilon=1$ and $\epsilon=3$, which are rep@1 and rep@3 respectively. From the table, it can be seen that our approach performs the best when the precision requirements are more strict.}}
	\begin{threeparttable}
		\begin{tabular}{ccccccc}
			\toprule
			\multirow{2}{*}{} & \multicolumn{2}{c}{HPatches \textit{illu}} & \multicolumn{2}{c}{HPatches \textit{view}}
			& \multicolumn{2}{c}{HPatches \textit{Overall}}\\ 
			& \%rep@1 & \%rep@3 & \%rep@1 & \%rep@3 & \%rep@1 & \%rep@3\\
			\midrule
			\multicolumn{1}{c}{SIFT}           &25.15  &43.89  &24.90  &52.16  &25.02  &48.10\\
			\multicolumn{1}{c}{SuperPoint}     &23.72  &50.45 &19.37  &51.00  &50.73  &21.50\\
			\multicolumn{1}{c}{LF-Net}       &26.68  &46.43  &17.45 &39.62  &21.99  &42.96\\
			\multicolumn{1}{c}{D2-Net}         &10.02  &36.43  &4.90  &31.08 &7.41 &33.71  \\
			\multicolumn{1}{c}{R2D2}       &31.32  &63.65  &27.46  &63.21  &29.36  &63.43\\
			\multicolumn{1}{c}{ASLFeat}           &32.77  &60.49  &28.32 &62.48  &30.51  &61.50\\
			\multicolumn{1}{c}{Key.Net}  &35.87  &59.19  &33.29  &62.94  &34.25  &60.59\\
			\midrule
			\multicolumn{1}{c}{$F$+GLE}         &\textbf{48.43}  &\textbf{61.64} &\textbf{42.84}  &\textbf{64.87}  &\textbf{45.58} &\textbf{64.28}\\
			\bottomrule
		\end{tabular}
	\end{threeparttable}
\end{table*}

\begin{table*}[htpb]
	\centering
	\caption{\small{The table compares our method to other methods including typical handcrafted features, self-supervised learning based ones and unsupervised leaning based ones via MMA. And the precision threshold is $\epsilon=3$. During comparison, we use HardNet++ to construct descriptors for all detectors and we can see that our approach performs the best.}}
	\begin{threeparttable}
		\begin{tabular}{ccccccc}
			\toprule
			\multirow{2}{*}{} & \multicolumn{2}{c}{HPatches \textit{illu}} & \multicolumn{2}{c}{HPatches \textit{view}}
			& \multicolumn{2}{c}{HPatches \textit{Overall}}\\ 
			& \%mma@1 & \%mma@3 & \%mma@1 & \%mma@3 & \%mma@1 & \%mma@3\\
			\midrule
			\multicolumn{1}{c}{SIFT}           &36.24  &47.92  &32.77  &51.77  &34.44 &49.92\\
			\multicolumn{1}{c}{SuperPoint}     &43.01  &69.38  &25.50  &59.88  &33.93 &64.46\\
			\multicolumn{1}{c}{LF-Net}       &37.87  &57.31  &24.02  &49.02   &30.69 &53.01\\
			\multicolumn{1}{c}{D2-Net}         &19.37  &50.21  &6.48  &31.62   &12.69 &40.57\\
			\multicolumn{1}{c}{R2D2}       &32.28  &71.47  &22.99  &66.00  &27.42 &68.64\\
			\multicolumn{1}{c}{ASLFeat}           &46.92  &77.45  &33.18 &67.44  &39.79 &72.26\\
			\multicolumn{1}{c}{Key.Net+HN++}  &43.32  &72.31  &33.24 &\textbf{70.52}  &38.09 &71.38\\
			\midrule
			\multicolumn{1}{c}{$F$+GLE+HN++}         &\textbf{64.24} &\textbf{79.27}  &\textbf{36.15}  &67.41 &\textbf{49.67}  &\textbf{73.12} \\
			\bottomrule
		\end{tabular}
	\end{threeparttable}
\end{table*}

\subsection{Repeatability}
In this part, we test the method on the entire Hpatches, where our methods are compared with the other three types of methods. The first type is of the handcrafted feature detection methods including Harris, FAST as well as SIFT. The second type is of supervised/self-supervised like LIFT and SuperPoint. The third one is of unsupervised learning-based pipelines including LF-Net, RF-Net, Key.Net, D2-Net and ASLFeat. We test the repeatability of different methods with error threshold $\epsilon$ of 1, 3 pixels respectively. Tab. 2 shows the average repeatability of each threshold on different validation set, including the HPatches with illumination changes, HPatches with viewpoint changes.

As shown in Tab. 2, our GLEO-Det has the best repeatability when $\epsilon=1$ and $\epsilon=3$, showinng that the GLEO-Det is stable for feature detection. Compared with D2-Net and ASLFeat, this demonstrates that our GLEO-Det can achieve fine constraint for the detector via local entropy taken into account. Moreover, we only train the detector using one epoch with 2000 iterations. This shows that the loss function based on local entropy can facilitate the training process of the detector, make it more efficient.

\begin{figure}[htpb]
	\centering
	\includegraphics[height=0.35\textwidth]{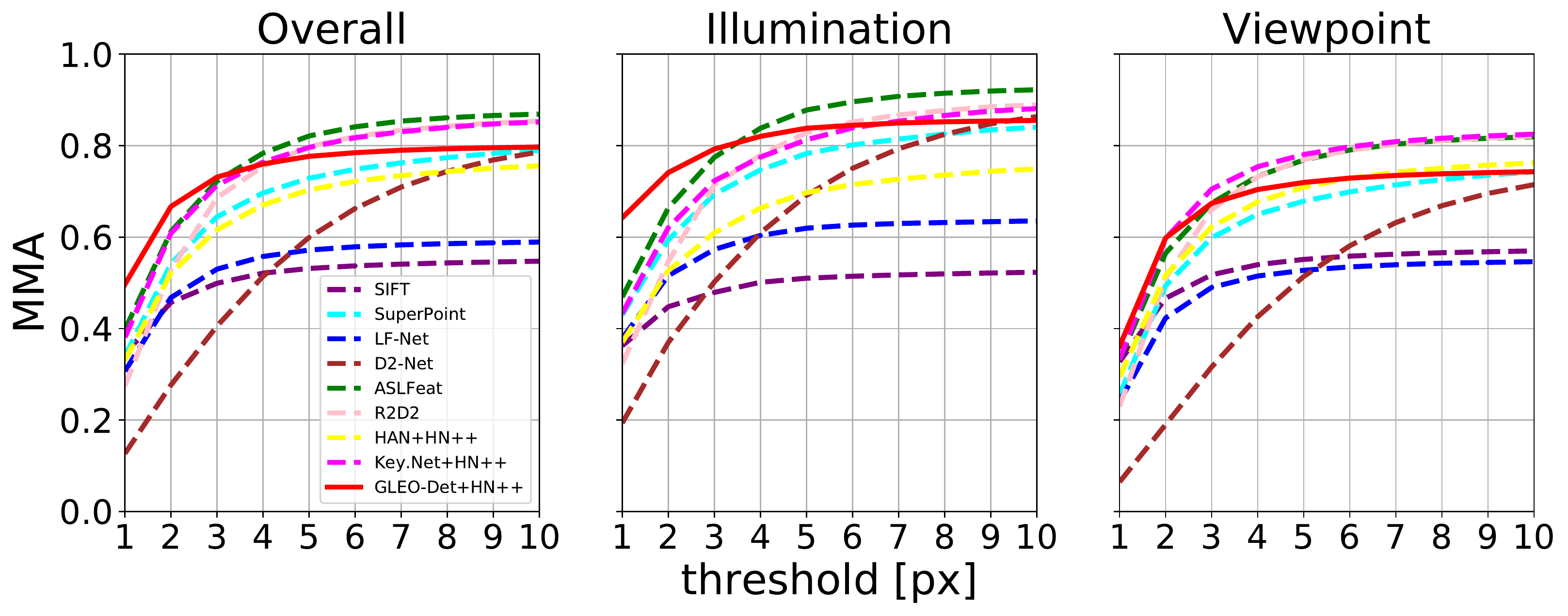}
	\caption{\small{The MMA curves of the methods in Tab. 3 under different thresholds. From the figure we can see that our approach shows the best performance when the threshold is small. However, it can also been seen that our method is worse than R2D2, ASLFeat and Key.Net when threshold is larger especially for viewpoint change case, which is because we haven't take scale factor into consideration and the performance of HN++ is limited.}}
	\label{Fig3.main}
\end{figure}

\subsection{Matchability}
For testing, we remove 8 groups in HPatches as in D2-Net. In this paper, we use the mean matching accuracy (MMA) to evaluate the feature detection algorithms. We also include the three types of approaches above for evaluation. In addition, for correspondence establishment, we use HardNet (HN++) \citep{mishchuk2017working} to build descriptors for the features and match through the nearest neighbor algorithm. The average MMA is shown in Tab. 3 and Fig. 3. 

From Tab. 3 and Fig. 3, we can see that our GLEO-Det shows competitive performance using HardNet descriptors for matching, especially when the threshold is more strict. And this tells that the features detected using GLEO-Det are not only precisely repeatable but also stable for matching. On the other hand, it can also be seen that our method performs worse than R2D2, ASLFeat and Key.Net when threshold is larger especially for viewpoint change case. We believe one reason is that we haven't taken scale factor into consideration as Key.Net and directly crop $32\times{32}$ patches surrounding key points for descriptor construction. The other reason is that the performance of HN++ is limited in certain degree.

\section{Conclusion}
In this paper, we propose an unsupervised pipeline for feature detection in images. Specifically, we attempt to combine the repeatability requirement and convolution feature guidance for better performance. In addition, entropy-based object function is developed to make it more easier to reach convergence. Finally, with the Guider block, we define the loss function from two sides, both positive and negative, in order to remove the points with little semantic information. However, the matching performance is still limited to the descriptors. Therefore, in our future work, we will attempt to establish guiding mechanism from detector to descriptor, and achieve mutual learning between both detector and descriptor.

\bibliographystyle{unsrtnat}
\bibliography{references}  

\begin{thebibliography}{26}
\providecommand{\natexlab}[1]{#1}
\providecommand{\url}[1]{\texttt{#1}}
\expandafter\ifx\csname urlstyle\endcsname\relax
  \providecommand{\doi}[1]{doi: #1}\else
  \providecommand{\doi}{doi: \begingroup \urlstyle{rm}\Url}\fi

\bibitem[Harris et~al.(1988)Harris, Stephens, et~al.]{harris1988combined}
Chris Harris, Mike Stephens, et~al.
\newblock A combined corner and edge detector.
\newblock In \emph{Alvey vision conference}, volume~15, pages 10--5244.
  Citeseer, 1988.

\bibitem[Smith and Brady(1997)]{smith1997susan}
Stephen~M Smith and J~Michael Brady.
\newblock Susan—a new approach to low level image processing.
\newblock \emph{International journal of computer vision}, 23\penalty0
  (1):\penalty0 45--78, 1997.

\bibitem[Trajkovi{\'c} and Hedley(1998)]{trajkovic1998fast}
Miroslav Trajkovi{\'c} and Mark Hedley.
\newblock Fast corner detection.
\newblock \emph{Image and vision computing}, 16\penalty0 (2):\penalty0 75--87,
  1998.

\bibitem[Rublee et~al.(2011)Rublee, Rabaud, Konolige, and
  Bradski]{rublee2011orb}
Ethan Rublee, Vincent Rabaud, Kurt Konolige, and Gary Bradski.
\newblock Orb: An efficient alternative to sift or surf.
\newblock In \emph{2011 International conference on computer vision}, pages
  2564--2571. Ieee, 2011.

\bibitem[Mair et~al.(2010)Mair, Hager, Burschka, Suppa, and
  Hirzinger]{mair2010adaptive}
Elmar Mair, Gregory~D Hager, Darius Burschka, Michael Suppa, and Gerhard
  Hirzinger.
\newblock Adaptive and generic corner detection based on the accelerated
  segment test.
\newblock In \emph{European conference on Computer vision}, pages 183--196.
  Springer, 2010.

\bibitem[Lindeberg(1998)]{lindeberg1998feature}
Tony Lindeberg.
\newblock Feature detection with automatic scale selection.
\newblock \emph{International journal of computer vision}, 30\penalty0
  (2):\penalty0 79--116, 1998.

\bibitem[Lowe(1999)]{lowe1999object}
David~G Lowe.
\newblock Object recognition from local scale-invariant features.
\newblock In \emph{Proceedings of the seventh IEEE international conference on
  computer vision}, volume~2, pages 1150--1157. Ieee, 1999.

\bibitem[Lowe(2004)]{lowe2004distinctive}
David~G Lowe.
\newblock Distinctive image features from scale-invariant keypoints.
\newblock \emph{International journal of computer vision}, 60\penalty0
  (2):\penalty0 91--110, 2004.

\bibitem[Bay et~al.(2006)Bay, Tuytelaars, and Gool]{bay2006surf}
Herbert Bay, Tinne Tuytelaars, and Luc~Van Gool.
\newblock Surf: Speeded up robust features.
\newblock In \emph{European conference on computer vision}, pages 404--417.
  Springer, 2006.

\bibitem[Matas et~al.(2004)Matas, Chum, Urban, and Pajdla]{matas2004robust}
Jiri Matas, Ondrej Chum, Martin Urban, and Tom{\'a}s Pajdla.
\newblock Robust wide-baseline stereo from maximally stable extremal regions.
\newblock \emph{Image and vision computing}, 22\penalty0 (10):\penalty0
  761--767, 2004.

\bibitem[LeCun et~al.(1998)LeCun, Bottou, Bengio, and
  Haffner]{lecun1998gradient}
Yann LeCun, L{\'e}on Bottou, Yoshua Bengio, and Patrick Haffner.
\newblock Gradient-based learning applied to document recognition.
\newblock \emph{Proceedings of the IEEE}, 86\penalty0 (11):\penalty0
  2278--2324, 1998.

\bibitem[Yi et~al.(2016)Yi, Trulls, Lepetit, and Fua]{yi2016lift}
Kwang~Moo Yi, Eduard Trulls, Vincent Lepetit, and Pascal Fua.
\newblock Lift: Learned invariant feature transform.
\newblock In \emph{European conference on computer vision}, pages 467--483.
  Springer, 2016.

\bibitem[Verdie et~al.(2015)Verdie, Yi, Fua, and Lepetit]{verdie2015tilde}
Yannick Verdie, Kwang Yi, Pascal Fua, and Vincent Lepetit.
\newblock Tilde: A temporally invariant learned detector.
\newblock In \emph{Proceedings of the IEEE conference on computer vision and
  pattern recognition}, pages 5279--5288, 2015.

\bibitem[DeTone et~al.(2018)DeTone, Malisiewicz, and
  Rabinovich]{detone2018superpoint}
Daniel DeTone, Tomasz Malisiewicz, and Andrew Rabinovich.
\newblock Superpoint: Self-supervised interest point detection and description.
\newblock In \emph{Proceedings of the IEEE conference on computer vision and
  pattern recognition workshops}, pages 224--236, 2018.

\bibitem[Ono et~al.(2018)Ono, Trulls, Fua, and Yi]{ono2018lf}
Yuki Ono, Eduard Trulls, Pascal Fua, and Kwang~Moo Yi.
\newblock Lf-net: Learning local features from images.
\newblock \emph{Advances in neural information processing systems}, 31, 2018.

\bibitem[Shen et~al.(2019)Shen, Wang, Li, Yu, Li, Wen, Cheng, and
  He]{shen2019rf}
Xuelun Shen, Cheng Wang, Xin Li, Zenglei Yu, Jonathan Li, Chenglu Wen, Ming
  Cheng, and Zijian He.
\newblock Rf-net: An end-to-end image matching network based on receptive
  field.
\newblock In \emph{Proceedings of the IEEE/CVF Conference on Computer Vision
  and Pattern Recognition}, pages 8132--8140, 2019.

\bibitem[Christiansen et~al.(2019)Christiansen, Kragh, Brodskiy, and
  Karstoft]{christiansen2019unsuperpoint}
Peter~Hviid Christiansen, Mikkel~Fly Kragh, Yury Brodskiy, and Henrik Karstoft.
\newblock Unsuperpoint: End-to-end unsupervised interest point detector and
  descriptor.
\newblock \emph{arXiv preprint arXiv:1907.04011}, 2019.

\bibitem[Barroso-Laguna et~al.(2019)Barroso-Laguna, Riba, Ponsa, and
  Mikolajczyk]{barroso2019key}
Axel Barroso-Laguna, Edgar Riba, Daniel Ponsa, and Krystian Mikolajczyk.
\newblock Key. net: Keypoint detection by handcrafted and learned cnn filters.
\newblock In \emph{Proceedings of the IEEE/CVF International Conference on
  Computer Vision}, pages 5836--5844, 2019.

\bibitem[Dusmanu et~al.(2019)Dusmanu, Rocco, Pajdla, Pollefeys, Sivic, Torii,
  and Sattler]{dusmanu2019d2}
Mihai Dusmanu, Ignacio Rocco, Tomas Pajdla, Marc Pollefeys, Josef Sivic,
  Akihiko Torii, and Torsten Sattler.
\newblock D2-net: A trainable cnn for joint description and detection of local
  features.
\newblock In \emph{Proceedings of the IEEE/cvf conference on computer vision
  and pattern recognition}, pages 8092--8101, 2019.

\bibitem[Revaud et~al.(2019)Revaud, Weinzaepfel, De~Souza, Pion, Csurka, Cabon,
  and Humenberger]{revaud2019r2d2}
Jerome Revaud, Philippe Weinzaepfel, C{\'e}sar De~Souza, Noe Pion, Gabriela
  Csurka, Yohann Cabon, and Martin Humenberger.
\newblock R2d2: repeatable and reliable detector and descriptor.
\newblock \emph{arXiv preprint arXiv:1906.06195}, 2019.

\bibitem[Luo et~al.(2020)Luo, Zhou, Bai, Chen, Zhang, Yao, Li, Fang, and
  Quan]{luo2020aslfeat}
Zixin Luo, Lei Zhou, Xuyang Bai, Hongkai Chen, Jiahui Zhang, Yao Yao, Shiwei
  Li, Tian Fang, and Long Quan.
\newblock Aslfeat: Learning local features of accurate shape and localization.
\newblock In \emph{Proceedings of the IEEE/CVF conference on computer vision
  and pattern recognition}, pages 6589--6598, 2020.

\bibitem[Simonyan and Zisserman(2014)]{simonyan2014very}
Karen Simonyan and Andrew Zisserman.
\newblock Very deep convolutional networks for large-scale image recognition.
\newblock \emph{arXiv preprint arXiv:1409.1556}, 2014.

\bibitem[Ronneberger et~al.(2015)Ronneberger, Fischer, and
  Brox]{ronneberger2015u}
Olaf Ronneberger, Philipp Fischer, and Thomas Brox.
\newblock U-net: Convolutional networks for biomedical image segmentation.
\newblock In \emph{International Conference on Medical image computing and
  computer-assisted intervention}, pages 234--241. Springer, 2015.

\bibitem[Ulyanov et~al.(2016)Ulyanov, Vedaldi, and
  Lempitsky]{ulyanov2016instance}
Dmitry Ulyanov, Andrea Vedaldi, and Victor Lempitsky.
\newblock Instance normalization: The missing ingredient for fast stylization.
\newblock \emph{arXiv preprint arXiv:1607.08022}, 2016.

\bibitem[Balntas et~al.(2017)Balntas, Lenc, Vedaldi, and
  Mikolajczyk]{balntas2017hpatches}
Vassileios Balntas, Karel Lenc, Andrea Vedaldi, and Krystian Mikolajczyk.
\newblock Hpatches: A benchmark and evaluation of handcrafted and learned local
  descriptors.
\newblock In \emph{Proceedings of the IEEE conference on computer vision and
  pattern recognition}, pages 5173--5182, 2017.

\bibitem[Mishchuk et~al.(2017)Mishchuk, Mishkin, Radenovic, and
  Matas]{mishchuk2017working}
Anastasiia Mishchuk, Dmytro Mishkin, Filip Radenovic, and Jiri Matas.
\newblock Working hard to know your neighbor's margins: Local descriptor
  learning loss.
\newblock \emph{Advances in neural information processing systems}, 30, 2017.

\end{thebibliography}






\end{document}